\pdfoutput=1
\documentclass[11pt]{article}

\usepackage[final]{EMNLP2025}

\definecolor{cvprblue}{rgb}{0.21,0.49,0.74}
\usepackage{amsmath}
\usepackage{times}
\usepackage{latexsym}
\usepackage[T1]{fontenc}
\usepackage[utf8]{inputenc}
\usepackage{microtype}
\usepackage{inconsolata}
\usepackage{xspace}
\usepackage{pifont}
\usepackage{multirow}
\usepackage{float}
\usepackage{dblfloatfix}
\usepackage{adjustbox}
\usepackage{lipsum} 
\usepackage[utf8]{inputenc} 
\usepackage[T1]{fontenc}    
\usepackage{url}            
\usepackage{booktabs}       
\usepackage{amsfonts}       
\usepackage{nicefrac}       
\usepackage{microtype}      
\usepackage{amssymb}
\usepackage{graphicx}
\usepackage{subcaption}
\usepackage{float}

\definecolor{Gray}{gray}{0.5}
\definecolor{LGray}{gray}{0.9}
\definecolor{darkblue}{RGB}{94,110,186}
\definecolor{darkGreen}{RGB}{92, 148, 110}
\definecolor{myblue}{RGB}{14, 121, 178}
\definecolor{myred}{RGB}{192, 0, 0}

\newcommand{\darkGreen}[1]{\textcolor{darkGreen}{#1}}

\newcommand{\ourmodel}{\text{LangDC}\xspace}
\newcommand{\ourmodelbold}{\textbf{{LangDC}}\xspace}
\newcommand{\ourmodule}{\text{CapPruner}\xspace}

\usepackage{soul}

\title{Seeing More, Saying More: Lightweight Language Experts are \\ Dynamic Video Token Compressors}

\author{
    Xiangchen Wang,$^{1}$\thanks{\ \ Equal contribution.} \quad 
    Jinrui Zhang,$^{1}$\footnotemark[1] \quad  
    Teng Wang,$^{1,2}$\footnotemark[1] \quad  \\
    \bf
    Haigang Zhang,$^3$ \ \ 
    Feng Zheng$^{1,4}$\thanks{\ \ Corresponding authors.} \\
    $^1$Southern University of Science and Technology \\
    $^2$The University of Hong Kong \\
    $^3$Shenzhen Polytechnic University 
    $^4$Spatialtemporal AI \\
    \small \texttt{\{wangxc2019, zhangjr2018\}@mail.sustech.edu.cn},   \texttt{ttengwang@gmail.com},   \texttt{zhengf@sustech.edu.cn} \\
}

\begin{document}
\maketitle 
\begin{abstract}

Recent advances in large video-language models have revolutionized video understanding tasks. However, their efficiency is greatly constrained by processing high volumes of visual tokens. Existing token compression strategies apply a fixed compression ratio, ignoring varying semantic density across video clips. Consequently, this leads to inadequate representation of information-rich clips due to insufficient tokens and unnecessary computation on static or content-poor ones. To address this, we propose \ourmodelbold, a \textbf{Lang}uage-aware \textbf{D}ynamic Token \textbf{C}ompressor. \ourmodel leverages a lightweight language model to describe video clips, converting them into soft caption tokens as visual representations. Trained with our proposed semantic density-aware supervision, \ourmodel aims to 1) cover key visual cues necessary for downstream task reasoning and 2) dynamically adjust compression ratios based on scene richness, reflected by description length. Our design mimics how humans dynamically express what they see: complex scenes (\textbf{seeing more}) elicit more detailed language to convey nuances (\textbf{saying more}), whereas simpler scenes are described with fewer words. Experimental results show that our method reduces FLOPs by 49\% compared to VideoGPT+ while maintaining competitive performance. Furthermore, qualitative results demonstrate our approach adaptively adjusts the token compression ratio based on video segment richness. Codes are available at \url{https://github.com/NIneeeeeem/LangDC}.

\end{abstract}    
\section{Introduction}
\label{sec:intro}

The field of video understanding has undergone a revolution thanks to recent advancements in large video-language models (LVLMs) ~\cite{liu2024visual, liu2024llavanext,2023videochat,chen2023minigptv2,Lin2023VideoLLaVALU,Luo2023ValleyVA}. By mapping visual token features to the embedding space of large language models (LLMs)~\cite{touvron2023llama, zheng2023judging, touvron2023llama2, chowdhery2023palm, chung2022scaling, ouyang2022training}, LVLMs provide a unified interface for video understanding tasks, enabling the capture of inter-task relationships and demonstrating exceptional generalization and reasoning capabilities. These breakthroughs pave the way for further progress in artificial general intelligence~\cite{Yu2019Lightweight, Guo2019Face}. However, the high computational cost of LVLMs, resulting from the quadratic complexity of processing numerous visual tokens with billion-scale parameters, impedes their real-world deployment. To alleviate this, considerable efforts have been made to derive compact, high-quality sets of visual tokens through carefully designed multimodal resamplers. These approaches include cross-attention-based methods (\textit{e.g.}, Q-Former~\cite{Li2023BLIP2BL,Ren2023TimeChatAT} and Resampler~\cite{Flamingo,Li2023LLaMAVIDAI,li2024tokenpacker}), convolution-based techniques (\textit{e.g.}, C-Abstractor~\cite{cha2023honeybee} and LDP~\cite{chu2023mobilevlm, chu2024mobilevlm}), and channel merging strategies such as pixel shuffle~\cite{Ren2023PixelLMPR,Chen2023InternVS} and adjacent concatenation~\cite{Bolya2022TokenMY,Song2023MovieChatFD}.

\begin{figure*}[htbp]  
    \centering         
    \includegraphics[width=0.99\textwidth]{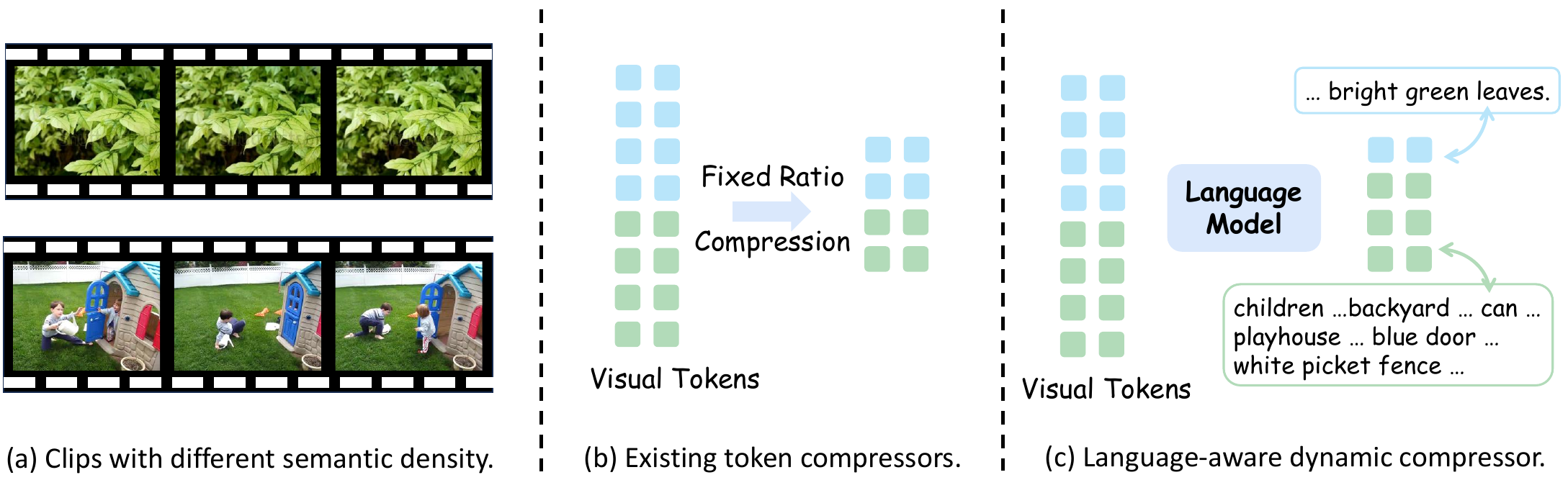}  
    \caption{\textbf{Comparison of \ourmodel and existing token compressors.} (a) illustrates two video segments with distinct information densities; the bottom segment contains richer visual cues. However, existing token compression methods (b) represent both segments to the same number of tokens. In contrast, our proposed method (c) dynamically allocates tokens based on semantic density, drawing on the sequence length awareness of language.}    
    \vspace{-6pt}
    \label{fig:comparision} 
\end{figure*}

While effective in improving efficiency, existing methods share a critical limitation: they apply a fixed compression ratio to visual tokens, disregarding variations in semantic density across video segments. For example, Figure~\ref{fig:comparision} (a) shows two clips with significantly different semantic densities: one is static, with each frame presenting close-ups of greenery, while the other is dynamic, showcasing various characters, objects, and actions. Despite this difference, both clips are compressed into the same number of visual tokens due to identical frame counts and resolutions. This uniform compression paradigm fails to produce an effective compact token set, as it may fail  to adequately represent information-rich segments while wasting tokens on less informative ones.

Inspired by the dynamic way of human language use in describing visual scenes, where simpler scenes are typically described with fewer words and information-rich scenes (``\textit{seeing more}'') require more detailed descriptions (``\textit{saying more}''), we propose \ourmodel, a language-aware dynamic token compressor. \ourmodel employs a lightweight language model to describe video segments, and then uses soft caption tokens (i.e., the hidden states of the predicted text tokens) as compressed visual representation. To ensure the compressed token set size reflects visual richness, we propose semantic density-aware supervision. Specifically, a strong LVLM~\cite{liu2024llavanext} extracts key visual cues from each segment, serving as targets for predictions of the lightweight language model. This explicit guidance enables \ourmodel to: 1) capitalize on the inherent correspondence between language length and semantic density, facilitating the dynamic control of token compression ratio, and 2) capture key visual clues that facilitating more compact representations and enhancing reasoning capabilities across diverse downstream tasks.

Experiments on diverse video understanding benchmarks validate our method's effectiveness and efficiency. Results show that \ourmodel reduces the FLOPs by 49\% while maintaining competitive performance compared to the strong baseline VideoGPT+~\cite{Maaz2024VideoGPT+}. This demonstrates that our method produces a more compact and semantically rich set of visual tokens. Additionally, \ourmodel outperforms existing state-of-the-art token compression techniques at similar compression ratios.  Qualitative results show that our approach adaptively adjusts the token compression ratio based on the scene richness of video segments.

To summarize, our contributions are threefold: 1) We propose \ourmodel, a novel language-aware token compression strategy. Using soft language tokens for visual representation, it adaptively adjusts compression ratios, improving token utilization over fixed-ratio techniques. 2) We propose semantic density-aware supervision for the token compressors. By explicitly providing reconstruction targets for token compression, we enable the derivation of a more compact feature set that is not only aware of information richness but also preserves key visual cues. 3) Experimental results demonstrate that our method reduces FLOPs by 49\% relative to the strong baseline VideoGPT+, while maintaining competitive performance. Additional qualitative results show adaptive compression based on video clip semantic density.

\section{Related Work}
\label{sec:related_work}
\begin{table*}[!ht]
    \centering
    \renewcommand{\arraystretch}{1.2}
    \setlength\tabcolsep{6pt}
    \resizebox{1.0\textwidth}{!}{
    \begin{tabular}{lcccccccc}
        \toprule
        \multirow{2}{*}{\textbf{Method}} & \multirow{2}{*}{\textbf{\# Tokens}$\downarrow$} & \multicolumn{6}{c}{\textbf{Sub-tasks}}  \\
         \cmidrule(lr){3-8}
        ~ & ~ & \textbf{Fine-grained Action} & \textbf{Object Existence} & \textbf{Moving Direction} & \textbf{Scene Transition} & \textbf{Moving Attribute} & \cellcolor{gray!20}\textbf{Avg.} \\
        \midrule
        Source of Video & -- & MiT V1 & CLEVRER &  CLEVRER &  MoVQA & CLEVRER & --\\
        
        \midrule
        AvgPooling $2\times2$ & 3328 & 47.0 & 81.0 & 37.0 & 38.5 & 85.5 & \cellcolor{gray!20}55.37 \\
        AvgPooling $4\times4$ & 832 & 44.0 & 73.5 & 26.5 & 36.5 & 78.0 & \cellcolor{gray!20}52.05 \\
        AvgPooling $8\times8$ & 208 & 48.0 & 67.0 & 26.0 & 40.5 & 59.0 & \cellcolor{gray!20}49.50 \\
        AvgPooling $16\times16$ & 80 & 44.0 & 49.5 & 19.5 & 38.0 & 49.0 & \cellcolor{gray!20}44.40 \\
        
        \midrule
         \textcolor{blue}{Oracle Performance} & \textcolor{blue}{--} & \textcolor{blue}{63.0} & \textcolor{blue}{96.5} & \textcolor{blue}{64.0} & \textcolor{blue}{91.0} & \textcolor{blue}{96.5} & \textcolor{blue}{72.4} \\
         \textcolor{blue}{Oracle Tokens} & \textcolor{blue}{--} & \textcolor{blue}{260.3} & \textcolor{blue}{274.3} & \textcolor{blue}{757.8} & \textcolor{blue}{156.5} & \textcolor{blue}{514.0} & \textcolor{blue}{354.48} \\

        \bottomrule
    \end{tabular}
    }
    \vspace{-2pt}
    \caption{
    \textbf{Performance comparison of LVLMs with varying compression ratios across multiple video understanding tasks.} Here, \textcolor{blue}{Oracle} denotes the ideal scenario where the highest compression ratio that yields the correct response is selected for each test instance. Our key observations are: (1) The ideal number of visual tokens varies significantly across different videos and tasks, and (2) an oracle model integrating multiple compression ratios consistently achieves superior performance.
    }
    \vspace{-6pt}
\label{tab:pool_analysis}
\end{table*}

\noindent\textbf{Large video-language models.} Large video-language models (LVLMs)~\cite{liu2024visual,2023videochat,chen2023minigptv2,Lin2023VideoLLaVALU,lin2023multi,Luo2023ValleyVA,Maaz2024VideoGPT+} have garnered significant attention recently. Leveraging large language models (LLMs)~\cite{touvron2023llama, zheng2023judging, chowdhery2023palm, chung2022scaling, ouyang2022training} as a unified task interface, LVLMs adapt to diverse video understanding tasks through flexible language instructions. Typically, an LVLM comprises three core components: a visual encoder to perceive frame-level information, a multimodal connector to align vision and language feature spaces, and an LLM for understanding and generating language content. Pretrained on large-scale visual-caption datasets and fine-tuned on video instruction data, LVLMs show superior performance over traditional task-specific models. Previous methods have enhanced LVLMs by: 1) collecting high-quality video instruction tuning data for versatile understanding~\cite{2023videochat, zhang2024reflective}, 2) utilizing stronger video encoders to capture fine-grained dynamics~\cite{li2023mvbench}, and 3) designing efficient connectors to improve efficiency~\cite{Li2023LLaMAVIDAI}. Our proposed method further improves multimodal connectors by enhancing flexibility through dynamic token customization based on visual information density in videos. 

\noindent\textbf{Visual token compressors.} 
Compressing visual tokens to enhance efficiency poses a crucial challenge in large vision-language models. Handling a substantial number of tokens produced by long-context visual inputs, such as videos and high-resolution images, using LLMs substantially escalates memory consumption and latency, thereby impeding real-world deployment. Various token compression techniques~\cite{chen2024image} have been proposed to shorten visual sequences. For instance, Q-Former and Resampler introduce a set number of trainable tokens that interact with visual features via cross-attention layers to capture essential visual cues~\cite{Li2023BLIP2BL,Ren2023TimeChatAT,Flamingo,Li2023LLaMAVIDAI,li2024tokenpacker}.
C-Abstractor and LDP downsample feature maps using convolutional layers, preserving spatial structure~\cite{cha2023honeybee, chu2024mobilevlm}. Other approaches directly apply simple channel-wise merging operations (e.g., mean-pooling, pixel-shuffle) following a multi-layer perceptron, effectively reducing model complexity while demonstrating strong generalization capabilities~\cite{Ren2023PixelLMPR,Chen2023InternVS,Bolya2022TokenMY,Song2023MovieChatFD}. Despite their effectiveness, these methods compress visual tokens using a fixed, predefined ratio, limiting their ability to generalize across samples with varying information density. In contrast, we utilize a pre-trained captioner to evaluate information density and generate soft caption tokens as compressed visual tokens, enabling adaptation to different visual inputs dynamically.
\begin{figure*}[htbp]  
    \centering         
    \includegraphics[width=\textwidth]{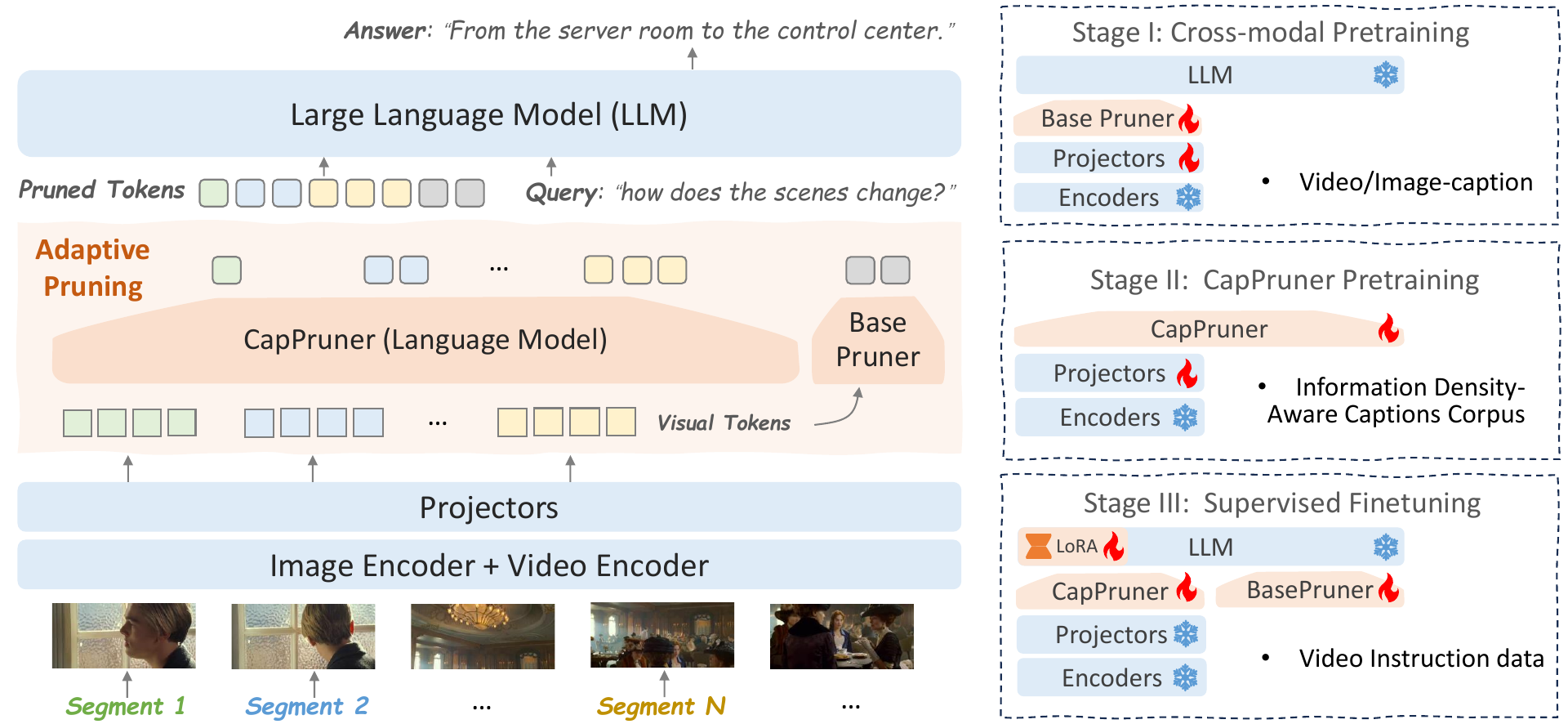}  
    \caption{\textbf{Overview of the proposed method.}
    \ourmodel utilizes dual visual encoders to extract visual features, followed by dynamic compression using CapPruner.
    The compressed features are combined with the base pruner’s output and fed into the LLM. The training pipeline consists of three stages: Stage I involves cross-modal pretraining with video/image-caption pairs, Stage II focuses on CapPruner pretraining using an information density-aware captions corpus, Stage III includes supervised fine-tuning with video instruction data.
    }
    \label{fig:model_architecture1} 
\end{figure*}

\section{Motivation on Dynamic Compression}

Intuitively, videos with varying information densities require different compression ratios. To validate this hypothesis, we conduct an in-depth analysis on five tasks of the MVBench~\cite{li2023mvbench}. Notably, this benchmark encompasses a wide range of subtasks and diverse data sources, and includes videos with distinct information densities—an attribute that makes it well-suited for our validation.

We train the MLLM~\cite{Maaz2024VideoGPT+} with different visual token compression ratio (implemented via  adaptive average pooling with different stride), and evaluate their optimal trade-off between token count and model performance. Specifically, we employ the oracle metric following~\cite{cai2024matryoshka},  which identifies the highest compression ratio that yields the correct response for each test instance, and subsequently compute both the token count and performance metrics.

As shown in Table~\ref{tab:pool_analysis}, higher compression ratios generally lead to reduced overall model performance. 
However, the non-uniform distribution of oracle token counts underscores the inherent variability of video information density, revealing the limitations of static token compression methods. 
Furthermore, the sensitivity of different task videos to changes in visual token counts varies significantly. For instance, in relatively static videos (e.g., State Changes from Prception Tests~\cite{Puatruaucean2023PerceptionTA}), decreasing the token count from $3k$ to $80$ results in only a $2\%$ drop in performance. Conversely, videos rich in elements and motion (such as those used in Moving Count task) experience a steep decline in accuracy as token counts decrease. These observations highlight the critical need for dynamic compression strategies adaptive to varying video content, suggesting this is the future direction for video compression.

\section{Methodology}

We propose \ourmodel, a \textbf{Lang}uage-aware \textbf{D}ynamic Token \textbf{C}ompressor, designed to dynamically compress visual content based on semantic richness. It is achieved through the integration of CapPruner, a lightweight language expert that transforms visual content into semantically rich token representations. Leveraging our proposed semantic density-aware supervision, CapPruner adaptively allocates the number of tokens according to the semantic density of the input. We start this section by first providing an overview of the \ourmodel's pipeline. Next, we detail the architecture and functionality of CapPruner and the semantic density-aware supervision mechanism. Finally, we outline the progressive training strategy employed for \ourmodel.

\paragraph{Overall architecture.} We build our model based on VideoGPT+~\cite{Maaz2024VideoGPT+}. As illustrated in Figure~\ref{fig:model_architecture1}, \ourmodel comprises dual visual encoders for spatial-temporal perception, a projector for vision-language feature alignment, token pruners for visual compression, and an LLM for language understanding and generation. The token pruner module incorporates a lightweight language expert, termed the dynamic token pruner (\textit{CapPruner}), alongside an adaptive mean pooler serving as the base pruner. Given an input video, we first divide it several segments and encode each seperately. The resulting features are subsequently passed through the projector and token pruners. The \textit{CapPruner} dynamically reduces the number of visual tokens within each segment, producing pruned tokens of variable lengths. These tokens are then temporally aggregated and combined with the output of the base pruner before being fed into the LLM for auto-regressive training or inference.

\subsection{Language-Aware Compression}

Dynamic compression hinges upon the effective capture of video semantics, which necessitating the integration of a pre-trained language model. However, departing from previous approaches~\cite{ye2024voco,shu2024video} that simply extract visual tokens, our method leverages the language expert to also determine the appropriate compression ratio. 
Therefore, language-aware dynamic token compressor capitalizes on the autoregressive nature of a language model, while simultaneously learning concise segment-level semantic representations from teacher model. This section details the training methodology and operational mechanism of the dynamic compressor.

\begin{figure}[t]  
    \centering         
    \includegraphics[width=0.98\linewidth]{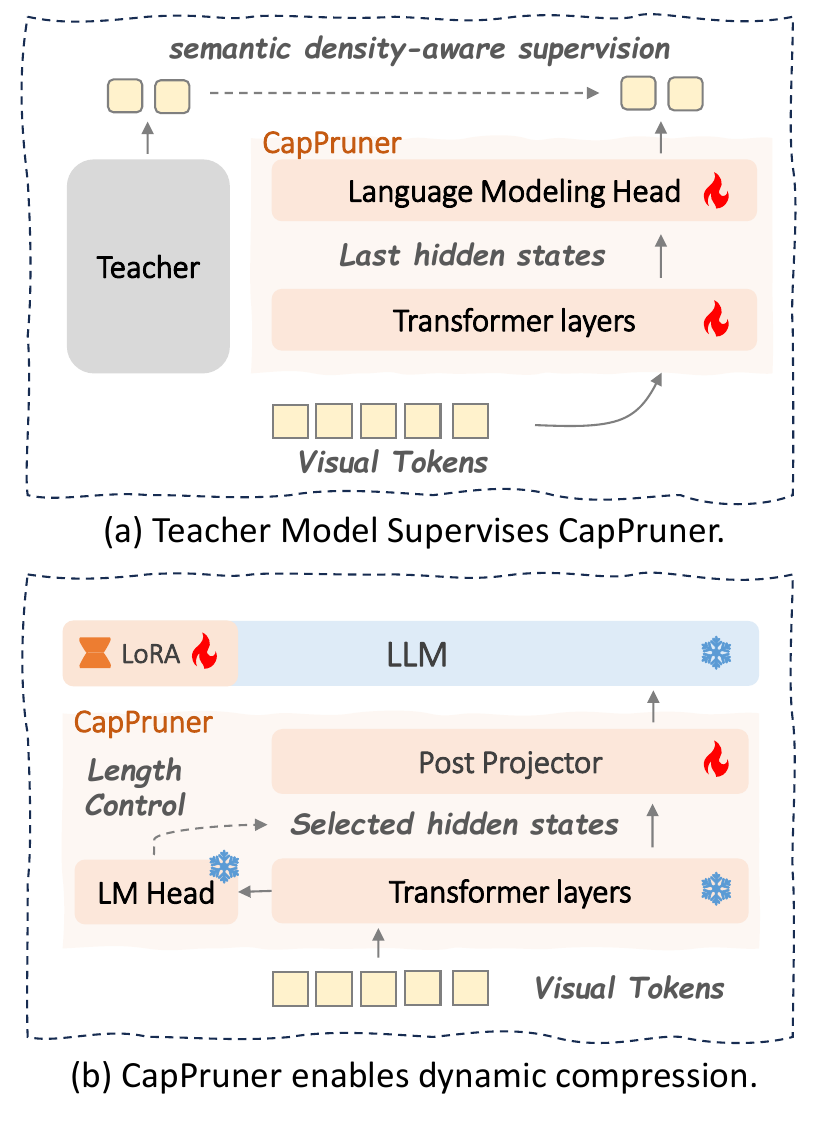}  
    \vspace{-6pt}
    \caption{
    \textbf{Illustration of the dynamic compression mechanism in CapPruner}. (a) Captions generated by a teacher model (a strong captioner) are used to supervise the training of CapPruner, facilitating it to allocate tokens according to scene richness. (b) By leveraging the hidden states of predicted captions as compact representation, CapPruner dynamically adjusts the compression ratio according to the timing of the “end-of-sentence” token prediction.
    }    
    \vspace{-6pt}
    \label{fig:cappruner} 
\end{figure}

\noindent\textbf{Captioner as pruner (CapPruner).} The CapPruner consists of a lightweight language model and two projection layers. In Figure~\ref{fig:cappruner}, the language model's transformer layers are utilized at various stages of training and inference to generate hidden states. The two projectors have distinct roles and are applied at different stages:

\begin{itemize}
    \item The \textit{language modeling head} from the lightweight language model serves as one projector. It maps the hidden state to the vocabulary, enabling supervised training based on important visual cues provided by a teacher model. This language modeling head is responsible for generating tokens and controlling their length. The "padding" token indicates that the compact visual representation are fully compressed.
    \item The other projector, known as the \textit{post projector}, aligns the dimensions of the hidden state with embeddings from the LLM, facilitating end-to-end instruction tuning and inference. Notably, CapPruner can select the optimal depth of hidden state for compressed visual features. In practice, hidden state from intermediate layers proves most effective, as shallower representations often lack sufficient semantic information, while deeper ones may exhibit excessive abstraction~\cite{Toneva2019InterpretingAI}. The detailed experimental results are provided in the supplementary materials.
\end{itemize}

\noindent\textbf{Semantic density-aware supervision.}
Effective visual semantic compression necessitates concise and dynamic supervision. Although manually annotated captions offer high accuracy, they are susceptible to annotator bias, resulting in discrepancies between caption length and the actual density of video information. Furthermore, manual annotations are resource-intensive, leading to limited dataset sizes and potential inconsistencies across datasets. To address these challenges, we leverage the consistent and descriptive capabilities of state-of-the-art vision-language models. Specifically, we employ LLaVA-OneVision~\cite{li2024llava} to extract crucial visual cues from each video segment.
By eliminating irrelevant and ambiguous language, we refine the supervisory signals to provide \ourmodule with a focused stream that accentuates essential visual information. This approach enhances the representation of core visual semantics, leading to more accurate compression results. The detailed processing procedure is demonstrated in the supplementary material. For a fair comparison with VideoGPT+~\cite{Maaz2024VideoGPT+}, teacher descriptions are constrained to video segments from the instruction tuning dataset. This practice preserves data consistency and isolates the influence of dynamic compression.

\begin{table*}
\centering
\label{tab:open-source-eval-results-wrt-duration}
\begin{adjustbox}{max width=\textwidth}
\begin{tabular}{l!{\vrule width \lightrulewidth}c!{\vrule width \lightrulewidth}c!{\vrule width \lightrulewidth}c!{\vrule width \lightrulewidth}c!{\vrule width \lightrulewidth}c!{\vrule width \lightrulewidth}c!{\vrule width \lightrulewidth}c} 
\toprule
\multicolumn{1}{l!{\vrule width \lightrulewidth}}{\multirow{2}{*}{\textbf{Models}}} & 
\multirow{2}{*}{\begin{tabular}[c]{@{}c@{}}\textbf{LLM}\\\textbf{\# Params}\end{tabular}} & 
\multicolumn{1}{c!{\vrule width \lightrulewidth}}{\multirow{2}{*}{\textbf{\# Frames}}} & 
\multirow{2}{*}{\begin{tabular}[c]{@{}c@{}}\textbf{SFT}\\\textbf{\# Pairs}\end{tabular}} & 
\multicolumn{2}{c!{\vrule width \lightrulewidth}}{\textbf{Video-MME}} & 
\multirow{2}{*}{\textbf{MVBench}} &\multirow{2}{*}{\begin{tabular}[c]{@{}c@{}}\textbf{Efficiency}\\\textbf{FLOPs$\downarrow$}\end{tabular}} \\
\cmidrule(lr){5-6}
                                              &        &     &                                                                        & w/o subs      & w/ subs                 &    \\  
\midrule

Video-LLaVA~\citep{videollava}                                                  & 7B         & 8 &       $765K$                                                                             & 39.9          & 41.6        &   -- & --      \\
ST-LLM~\citep{st-llm}                                                         & 7B      & 64 &                                                         $330K$                              & 37.9          & 42.3       & 54.8 & --              \\
VideoChat2 \citep{li2023mvbench}                                          & 7B            & 16 &     $2M$                                                                                & 39.5          & 43.8   &     51.1 & --             \\
Chat-UniVi-V1.5~\citep{jin2023chatunivi}                                          & 7B       & 64 &   $649K$                                                                                         & 40.6          & 45.9       &  -- & --              \\

VideoGPT+~\citep{Maaz2024VideoGPT+}  & 3.8B             & 16 &             $330K$                                                    & \textbf{44.5}          &   \underline{49.9}       &     \textbf{58.7}   & 49.85T      \\
\rowcolor[HTML]{deebf7} \ourmodel~(ours)  & 3B                & 16 &          $330K$                                                                   & \underline{44.3}          & \textbf{51.3}     & \underline{57.1}      &     25.15T     \\

\bottomrule

\end{tabular}
\end{adjustbox}
\caption{\textbf{Performance comparison with baselines on Video-MME and MVBench.}}
\label{tab:videomme}
\end{table*}

\subsection{Training Recipe}

Traditional practices for LVLMs suggest that a progressive training strategy is essential to reduce the semantic gap between visual and linguistic representations. Our proposed method, \ourmodel, incorporates a lightweight language expert with built-in knowledge of the semantic space. This expert module is crucial for establishing links between visual representations and language embeddings, requiring a distinctive progressive training approach that aligns spatial representations across different modalities. The training process comprises three sequential stages (shown in Fig.~\ref{fig:model_architecture1}): 

\noindent\textbf{Cross-modal pretraining.}
The pretraining phase aims to establish alignment between visual and textual representations. Following \cite{liu2024visual}, the projectors connecting the visual encoders to both the CapPruner and the LLM are trained, while all other model components remain frozen.

\noindent\textbf{CapPruner pretraining.} We first train CapPruner with a base caption dataset to enable it to capture the fine-grained details of visual content. To further ensure that \ourmodule follows the principle of "seeing more, saying more", further refinement is required. As explained in the previous section, a state-of-the-art LVLM assists the lightweight language expert in producing descriptions of variable lengths that match the information density of the video segments. During this training phase, both \ourmodule and the associated visual encoder projectors are engaged, using the generated captions as supervision signals. Subsequently, \ourmodule is linked to the base LLM through a post-projector, which is initialized by the same data with the cross-modal pretraining stage.

\noindent\textbf{Supervised finetuning.}
During supervised finetuning, the model is trained to understand human instructions. The LoRA method with a rank of 128 is implemented on LLM. The interconnecting projectors between the language expert and LLM are fully trained, while all other components are frozen. Furthermore, the Adapt Token Pruner utilizes a teacher forcing mechanism to improve training efficiency during this stage.

\section{Experiments}

\subsection{Experiments Setup}

\noindent\textbf{Implementation details}. Following VideoGPT+, we adopt a dual-encoder setup comprising an image encoder (CLIP-ViT-L/14-336~\cite{radford2021clip}) and a video encoder (InternVideo2-stage-2-1B~\cite{wang2024internvideo2}). Unless otherwise noted, we apply  $4\times4$ pooling as the BasePruner, initialize the CapPruner with Qwen-2.5-0.5B and employ Qwen-2.5-3B~\cite{qwen2.5} for the LLM.
For cross-modal pre-training, the CC-595K dataset~\cite{liu2024visual} is used to independently train the image and video projectors. 
Supervised fine-tuning follows the procedure in VideoGPT+~\cite{Maaz2024VideoGPT+}, leveraging two instruction-tuning datasets tailored for distinct task formats. Additional details are provided in the supplementary material.

\begin{table*}[!ht]
\centering
\resizebox{0.92\textwidth}{!}{
\begin{tabular}{l c c c c c c c c c c c}
\toprule
\multirow{2}{*}{\textbf{Models}} & \textbf{LLM} & \multicolumn{2}{c}{\textbf{MSVD-QA}} & \multicolumn{2}{c}{\textbf{MSRVTT-QA}} & \multicolumn{2}{c}{\textbf{TGIF-QA}} & \multicolumn{2}{c}{\textbf{ActivityNet-QA}} \\
\cmidrule{3-10}
 & \textbf{\# Params} & \textbf{Accuracy} & \textbf{Score} & \textbf{Accuracy} & \textbf{Score} & \textbf{Accuracy} & \textbf{Score} & \textbf{Accuracy} & \textbf{Score}  \\
\midrule
VideoChat~\cite{2023videochat} & 7B & 56.3 & 2.8 & 45.0 & 2.5 & 34.4 & 2.3 & 26.5 & 2.2 \\
LLaMA Adapter~\cite{llama_adapter} & 7B & 54.9 & 3.1 & 43.8 & 2.7 & - & - & 34.2 & 2.7 \\
Video-LLaMA~\cite{damonlpsg2023videollama} & 7B & 51.6 & 2.5 & 29.6 & 1.8 & - & - & 12.4 & 1.1 \\
Video-ChatGPT~\cite{Maaz2023VideoChatGPT} & 7B & 64.9 & 3.3 & 49.3 & 2.8 & 51.4 & 3.0 & 35.2 & 2.8 \\
ChatUniVi~\cite{jin2023chatunivi} & 7B & 65.0 & 3.6 & 54.6 & 3.1 & 60.3 & 3.4 & 45.8 & 3.2 \\
LLaMA-VID~\cite{Li2023LLaMAVIDAI} & 7B & 70.0 & 3.7 & 58.9 & 3.3 & -- & -- & 47.5 & 3.3 \\
Video-LLaVA~\cite{Lin2023VideoLLaVALU} & 7B & 70.7 & \underline{3.9} & 59.2 & \underline{3.5} & 70.0 & 4.0 & 45.3 & 3.3 \\
VideChat2~\cite{li2023mvbench} & 7B & 70.0 & \underline{3.9} & 54.1 & 3.3 & -- & -- & 49.1 & 3.3 \\
VideoGPT+~\cite{Maaz2024VideoGPT+} & 3.8B & \underline{72.4} & \underline{3.9} & \textbf{60.6} & \textbf{3.6} & \underline{74.6} & \underline{4.1} & \underline{50.6} & \textbf{3.6} \\
 LongVLM~\cite{LongVLM} & 7B & 70.0 & 3.8 & 59.8 & 3.3 & -- & -- & 47.6 & 3.3 \\
 LLAVA-Mini~\cite{llavamini} & 7B & 70.9 & 4.0 & 59.5 & \textbf{3.6} & -- & -- & \textbf{53.5} & \underline{3.5} \\

 \rowcolor[HTML]{deebf7} \ourmodel~(ours) & \textbf{3B} & \textbf{74.0}& \textbf{4.0}  & \underline{59.9} & \textbf{3.6} & \textbf{76.8} & \textbf{4.2}  & 50.3 & \underline{3.5} \\
\bottomrule
\end{tabular}
}
\caption{\textbf{Performance comparison with baselines on four open-ended VideoQA benchmarks.}}

\label{tab:zeroshot}
\end{table*}

\begin{table*}[!ht]
    \centering
    \renewcommand{\arraystretch}{1.2}
    \setlength\tabcolsep{3pt}
    \resizebox{1.0\textwidth}{!}{
    \begin{tabular}{lcccccccccccccccccccccc}
        \toprule
        \multirow{2}{*}{\textbf{Models}} & \multicolumn{21}{c}{\textbf{Reference Metrics}} & {\textbf{Efficiency}} \\
        \cmidrule(lr){2-22} \cmidrule(lr){23-23}
        & \textbf{AS} & \textbf{AP} & \textbf{AA} & \textbf{FA} & \textbf{UA} & \textbf{OE} & \textbf{OI} & \textbf{OS} & \textbf{MD} & \textbf{AL} & \textbf{ST} & \textbf{AC} & \textbf{MC} & \textbf{MA} & \textbf{SC} & \textbf{FP} & \textbf{CO} & \textbf{EN} & \textbf{ER} & \textbf{CI} & \cellcolor{gray!20}\textbf{Avg.} & \# Tokens$\downarrow$ \\
        \midrule
        AvgPooling $2\times2$ &  72.5 & 57.5 & 88.9 & 47.0 & 59.0 & 81.0 & 75.0 & 35.5 & 37.0 & 34.5 & 86.0 & 38.5 & 65.0 & 85.5 & 41.0 & 41.8 & 49.5 & 33.0 & 42.0 & 57.5 & \cellcolor{gray!20}{55.37} & 3328  \\
AvgPooling $4\times4$ &  67.5 & 54.0 & 73.7 & 44.0 & 57.0 & 73.5 & 70.5 & 35.0 & 26.5 & 35.0 & 85.5 & 36.5 & 54.5 & 78.0 & 40.0 & 40.5 & 43.0 & 34.0 & 40.0 & 52.5 & \cellcolor{gray!20}{52.05} & 832 \\
AvgPooling $8\times8$ & 66.0 & 52.5 & 76.8 & 48.0 & 53.5 & 67.0 & 69.5 & 40.0 & 26.0 & 34.0 & 79.0 & 40.5 & 50.0 & 59.0 & 39.5 & 37.0 & 38.5 & 33.5 & 36.0 & 44.0 & \cellcolor{gray!20}{49.50} & 208 \\
AvgPooling $16\times16$ & 57.5 & 45.0 & 69.7 & 44.0 & 49.5 & 49.5 & 68.5 & 33.0 & 19.5 & 28.0 & 80.0 & 38.0 & 47.0 & 49.0 & 39.0 & 34.5 & 33.0 & 32.0 & 35.5 & 36.0 & \cellcolor{gray!20}{44.40} & 80 \\
\rowcolor[HTML]{deebf7} \ourmodel (w/ AvgPooling) & 68.5 & 51.5 & 88.5 & 49.5 & 57.0 & 79.5 & 65.5 & 34.0 & 37.5 & 31.5 & 87.5 & 42.5 & 67.0 & 76.5 & 41.0 & 39.5 & 47.5 & 30.5 & 39.5 & 56.0 & 54.52 & 1068$^{\dag}$ \\
\midrule
LDPv2~\cite{chu2024mobilevlm} &65.5&56&82.3&45.5&57.5&69.0&68.5& 36.5 & 25.0 & 32.5 & 83.0 & 39.5 & 51.5 & 61.5 & 37.5 & 36.5 & 37.5 & 32.5 & 38.5 & 50.5 & \cellcolor{gray!20}{50.29} & 512\\
        LDPv2~\cite{chu2024mobilevlm} & 71.0&54.5&84.8&48.0&58.0&79.5&75.5&35.5&31.5&34.5&82.0&43.5&59.5&79.5&39.0&42.0&36.5&33.5&36.5&57.0 & \cellcolor{gray!20}{54.08} & 1136 \\
        Resampler & 67.0 & 51.5 & 79.8 & 43.5 & 54.0 & 62.0 & 70.5 & 29.0 & 26.0 & 30.5 & 85.0 & 46.0 & 49.5 & 54.0 & 42.0 & 40.0 & 38.5 & 31.5 & 35.0 & 45.0 & \cellcolor{gray!20}{49.0} & 832 \\
        C-Abstractor~\cite{cha2023honeybee} & 69.5 & 57.5 & 84.3 & 45.5 & 59.0 & 79.5 & 69.0 & 33.5 & 31.0 & 34.5 & 85.5 & 46.0 & 59.0 & 74.5 & 36.5 & 39.0 & 37.0 & 37.0 & 38.0 & 54.5 & \cellcolor{gray!20}{53.5} & 832 \\
       \rowcolor[HTML]{deebf7} \ourmodel (w/ LDPv2) & 66.0 & 55.5 & 86.0 & 46.5 & 57.0 & 74.0 & 72.0 & 37.5 & 36.5 & 35.0 & 86.5 & 43.5 & 63.0 & 74.0 & 40.5 & 40.0 & 44.5 & 33.0 & 40.0 & 51.5 & 54.13 & 748$^{\dag}$ \\
        \bottomrule
    \end{tabular}
    }
    \caption{\textbf{Performance comparison of different token compressors on MVBench.} w/ LDPv2 means LDPv2 is utilized as base pruner. \dag ~indicates that the number of tokens varies across different test instances; we report the average value across all samples.}
\label{tab:compression_compare}
\end{table*}

\noindent\textbf{Evaluation benchmarks.} We evaluate \ourmodel on both multiple-choice and open-ended VideoQA tasks.
For multiple-choice benchmarks, we use MVBench~\cite{li2023mvbench} and VideoMME~\cite{Fu2024VideoMMETF}.
For open-ended VideoQA, we evaluate our model on MSVD-QA~\cite{msvd}, MSRVTT-QA, ActivityNet-QA and TGIF-QA~\cite{TGIF}. Following prior work~\cite{Maaz2024VideoGPT+}, we utilize GPT-3.5-Turbo-0613 to assess response accuracy, with scoring prompts detailed in the supplementary material.

\subsection{Main Results}

\noindent\textbf{Performance evaluation.}
Table~\ref{tab:videomme} shows that \ourmodel outperforms state-of-the-art LVLMs while reducing computational costs. Compared to VideoGPT +, \ourmodel reduces TFLOPs by 49\% with only a performance drop of 1.6\% on MVBench. This highlights the efficiency of semantic density-aware supervision in preserving key visual information.  On Video-MME, \ourmodel achieves superior performance with fewer parameters and less fine-tuning data. Notably, it drops only 0.2\% without subtitles and exceeds VideoGPT+ by 1.4\% with subtitles, excelling especially on long-video tasks which demonstrating CapPruner’s strength in long-range understanding.

Table~\ref{tab:zeroshot} shows that \ourmodel also surpasses VideoGPT+ by $1.6\%$ on MSVD-QA and $2.2\%$ on TGIF-QA, while remaining competitive on MSRVTT-QA and ActivityNet-QA. 
These results confirm CapPruner’s dynamic compression improves efficiency and preserves key semantic details, boosting generalization in zero-shot settings.

\begin{figure}[tbp]  
    \centering         
    \includegraphics[width=0.98\linewidth]{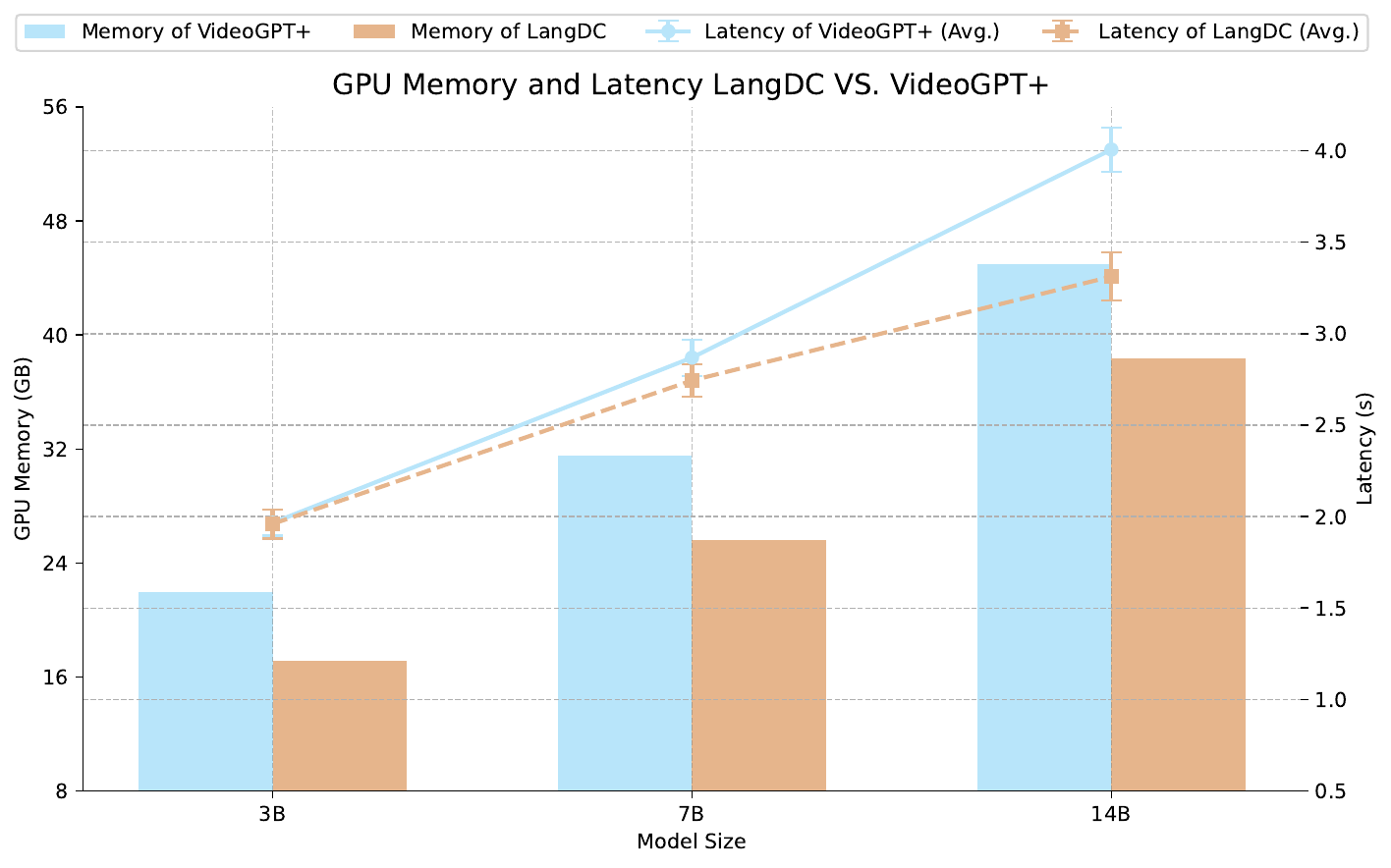}  
    \caption{\textbf{Comparison of GPU Memory and Latency.}} 
    \label{fig:inference} 
\end{figure}

\noindent\textbf{Efficiency analysis.} 
LangDC compress visual tokens from $3328$ to approximately $1068$, reducing computational cost from $49.85$ TFLOPs to $25.15$ TFLOPs. As shown in Figure\ref{fig:inference}, it also reduces GPU memory and latency compared to pooling, even with an added lightweight LLM. Notably, \ourmodel’s efficiency gains scale with larger base LLMs.
And table \ref{tab:compression_compare} further compares \ourmodel with other compression methods. Compared to the naive pooling compression strategy, LangDC matches the performance of a solution that uses three times as many tokens, and surpasses carefully designed compression modules like LDPv2\cite{chu2024mobilevlm}. 
Replacing BasePruner with LDPv2 further improves efficiency, surpassing C-Abstractor and Resampler by 0.6 and 5.1 points while requiring 100 fewer tokens. For fairness, all methods use the same pretraining and tuning data.

\begin{figure*}[htbp]  
    \centering         
    \includegraphics[width=0.98\textwidth]{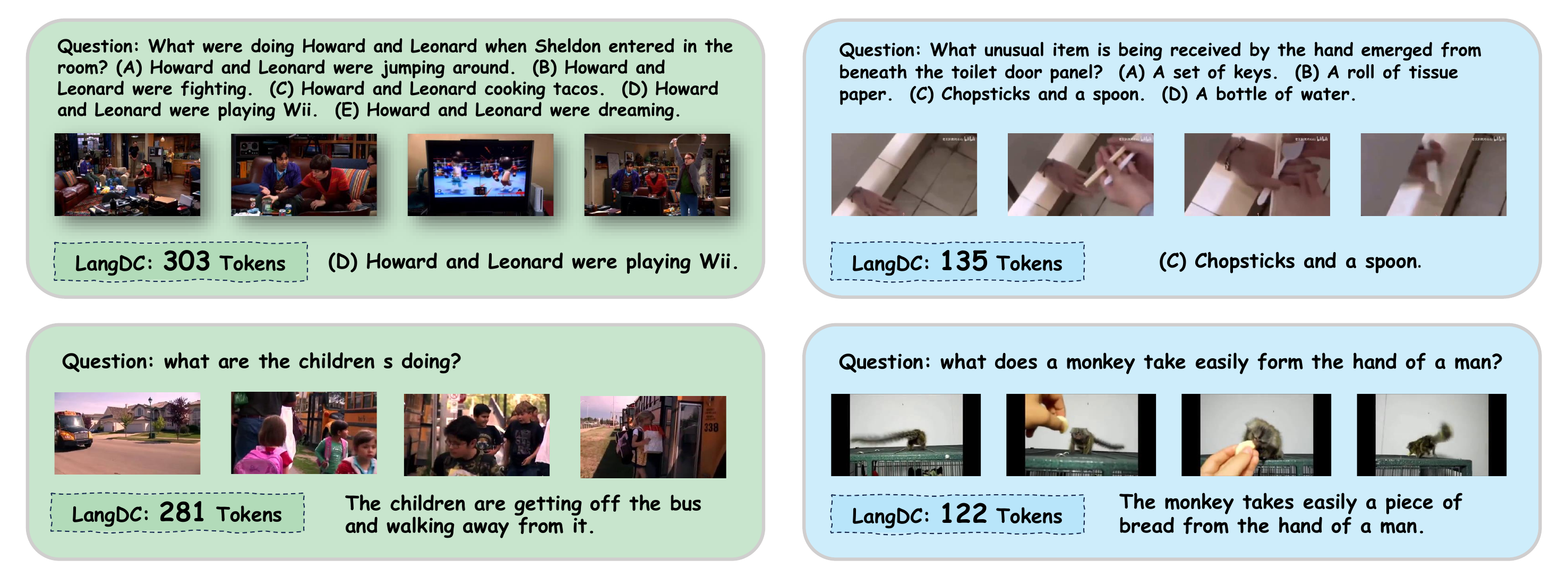}  
    \caption{\textbf{Visualization of video QA examples alongside the corresponding number of allocated tokens.}} 
    \label{fig:visual} 
\end{figure*}

\subsection{Ablation Studies}

This section provides a comprehensive analysis of CapPruner, exploring its dynamic characteristics, training schemes, supervision signals and pruner combinations. Qwen2.5-1.5B serves as the LLM.

\noindent\textbf{Dynamic vs. fixed compression ratio.} 
To highlight the strength of dynamic compression, we complement  qualitative results in Figure~\ref{fig:visual}, showing that CapPruner allocates more tokens to visually rich or action-intensive videos, and fewer to simpler ones. Table~\ref{tab:re_tokennum} further confirms its ability.

\begin{table}[!t]
\centering
\setlength{\tabcolsep}{3pt}
\renewcommand{\arraystretch}{0.7}
\footnotesize
\resizebox{\linewidth}{!}{
\begin{tabular}{c c | c c }
\toprule
Action Antonym & Object Existence   & State Change & Episodic Reasoning   \\
\midrule
143.2 & 184.7 & 249.1  & 257.2    \\ 
\bottomrule
\end{tabular}
}
\caption{\textbf{Comparison of exact token numbers of \ourmodel across diverse tasks within MVBench.}}
\label{tab:re_tokennum}
\end{table}

\begin{table}[!t]
    \centering
    \setlength\tabcolsep{6.0pt}
    \renewcommand{\arraystretch}{1.2}
    \resizebox{1.0\linewidth}{!}{
        \begin{tabular}{cc|c|cc}
        \toprule
         \textbf{BasePruner} & \textbf{CapPruner} & \textbf{Accuracy} & \textbf{\# Tokens} & \textbf{FLOPs}\\
        \midrule
         \ding{55} & \ding{52} & 51.50 & 236$^{\textcolor{red}{\dag}}$ & 18.24T \\
           AvgPooling $8\times8$ & \ding{55} & 49.50 & 208 & 16.06T \\
        AvgPooling $8\times8$ & \ding{52} & 51.62 & 444$^{\textcolor{red}{\dag}}$ & 19.51T \\
         AvgPooling $4\times4$ & \ding{55} & 52.05 & 832 & 17.57T \\

        AvgPooling $4\times4$ & \ding{52} & 54.52 & 1068$^{\textcolor{red}{\dag}}$ & 21.38T \\
        
        \bottomrule
        \end{tabular}
    }
    \caption{
    \textbf{Ablation of the combinations of BasePruner and CapPruner on MVBench}. \textcolor{red}{\dag} indicates that the \# tokens is not fixed. 
    }
    \label{tab:ablation_branch}
\end{table}


\noindent\textbf{Ablation of different pruners.} Table \ref{tab:ablation_branch} reports ablation results on MVBench with different combinations of CapPruner and BasePruner.
Using CapPruner alone yields 51.50\% accuracy with 236 tokens. In comparison, BasePruner with $8\times8$ pooling achieved lower accuracy of $49.50\%$ with a similar token number, while $4\times4$ pooling achieved a slightly higher but at the cost of significantly more tokens.
Importantly, combining CapPruner with either pooling strategy consistently improves accuracy.
Furthermore, CapPruner is compatible with other compressors: as shown in Table~\ref{tab:compression_compare}, pairing it with LDPv2 yields substantial performance gains.

\noindent\textbf{Ablation of the training scheme.}
Table~\ref{tab:re_schemes} demonstrates the critical role of CapPruner pretraining, which improves average accuracy from 45.40\% to 54.52\%. Post-pretraining further strengthens the integration between CapPruner and the LLM, leading to an additional increase from 49.12\% to 54.52\%.

\begin{table}[!t]
\centering
\setlength{\tabcolsep}{8pt}
\renewcommand{\arraystretch}{0.9}
\small
\begin{tabular}{l c}
\toprule

\textbf{Training Schemes} & \textbf{Accuracy}  \\
\midrule
Full CapPruner Pretraining &  \textbf{54.52} \\ 
w/o Post-Pretraining &  49.12 \\ 
w/o CapPruner-Pretraining &  45.40 \\
\bottomrule
\end{tabular}
\caption{\textbf{Ablation of the training scheme on MVBench.}}
\label{tab:re_schemes}
\end{table}

\begin{table}[!t]
\centering
\setlength{\tabcolsep}{3pt}
\renewcommand{\arraystretch}{0.8}
\footnotesize
\resizebox{1.0\linewidth}{!}{
\begin{tabular}{c | c c c}
\toprule
Method  & Pooling $2\times 2^{\textcolor{red}{\dag}}$ & Pooling $4\times 4$ & LangDC   \\
\midrule
 w/o captions &  55.37 & 52.05 & 54.52 \\
 w/ caption &  55.63 ({$\uparrow$0.26}) & 52.32 ({$\uparrow$0.27}) & 54.66 ({$\uparrow$0.14}) \\ 

\bottomrule
\end{tabular}
}
\caption{\textbf{Impact of caption supervision signal.} $^{\textcolor{red}{\dag}}$ indicates the same compression strategy as VideoGPT+.}
\label{tab:re_fairdata}
\end{table}

\noindent\textbf{Impact of caption supervision signal.}
Table \ref{tab:re_fairdata} highlights the effect of caption supervision signals in LangDC, particularly for regulating caption length. While incorporating it during pretraining yields only a modest improvement, the results suggest its overall influence on pretraining is limited.

\section{Discussion and Conclusion}

This study introduced LangDC, a language-aware dynamic token compressor for video understanding. Addressing the limitations of fixed compression ratios, which often fail to capture the varying semantic density of video content, LangDC leverages CapPruner to generate soft caption tokens as compressed visual representations. 
Guided by semantic-aware supervision, it effectively captures key visual cues while adjusting compression dynamically.
Extensive experiments across benchmarks with varying semantic densities demonstrate the superior performance-computation trade-off offered by LangDC’s adaptive token allocation. 
This strategy not only enhances efficiency but also sets a foundation for future research into more sophisticated, adaptive video understanding methods.
\section*{Limitations}

While our dynamic compression mechanism demonstrates human-aligned linguistic patterns and significantly enhances computational efficiency, two critical limitations warrant attention. First, given current resource constraints, our experiments focus on 1.5B/3B LLM configurations, leaving open questions about architectural scaling effects. Second, though the visual density-optimized compression strategy shows strong multi-turn dialog compatibility, its single-ratio implementation may partially constrain adaptability for specialized video QA tasks.
\section*{Acknowledgements}
This work was supported by the National Key R\&D Program of China (2024YFE0203100) and the Center for Computational Science and Engineering atSouthern University of Science and Technology.
\bibliography{main}
\appendix
\clearpage
\setcounter{page}{1}

\renewcommand\thefigure{\Alph{section}\arabic{figure}}
\renewcommand\thetable{\Alph{section}\arabic{table}}
\setcounter{figure}{0}
\setcounter{table}{0}

\section{Additional Results}

\noindent\textbf{Comparison of downsampling rates for pooling.}
\label{sec:pool_analysis}
Tab~\ref{tab:full_pool_analysis} confirms that different videos contain varying information densities, necessitating different token counts. 
We tested all subtasks of MVBench with pooling strategies of varying compression rates and calculated the \textbf{Oracle}, the scenario where the best tradeoff between visual tokens and performance is selected. The optimal number of tokens fluctuates across different videos and tasks and the oracle model integrates multiple pooling strategies achieves superior performance.

\begin{figure}[th]
    \centering
    \begin{subfigure}[b]{\columnwidth}
        \centering
        \includegraphics[width=0.96\textwidth]{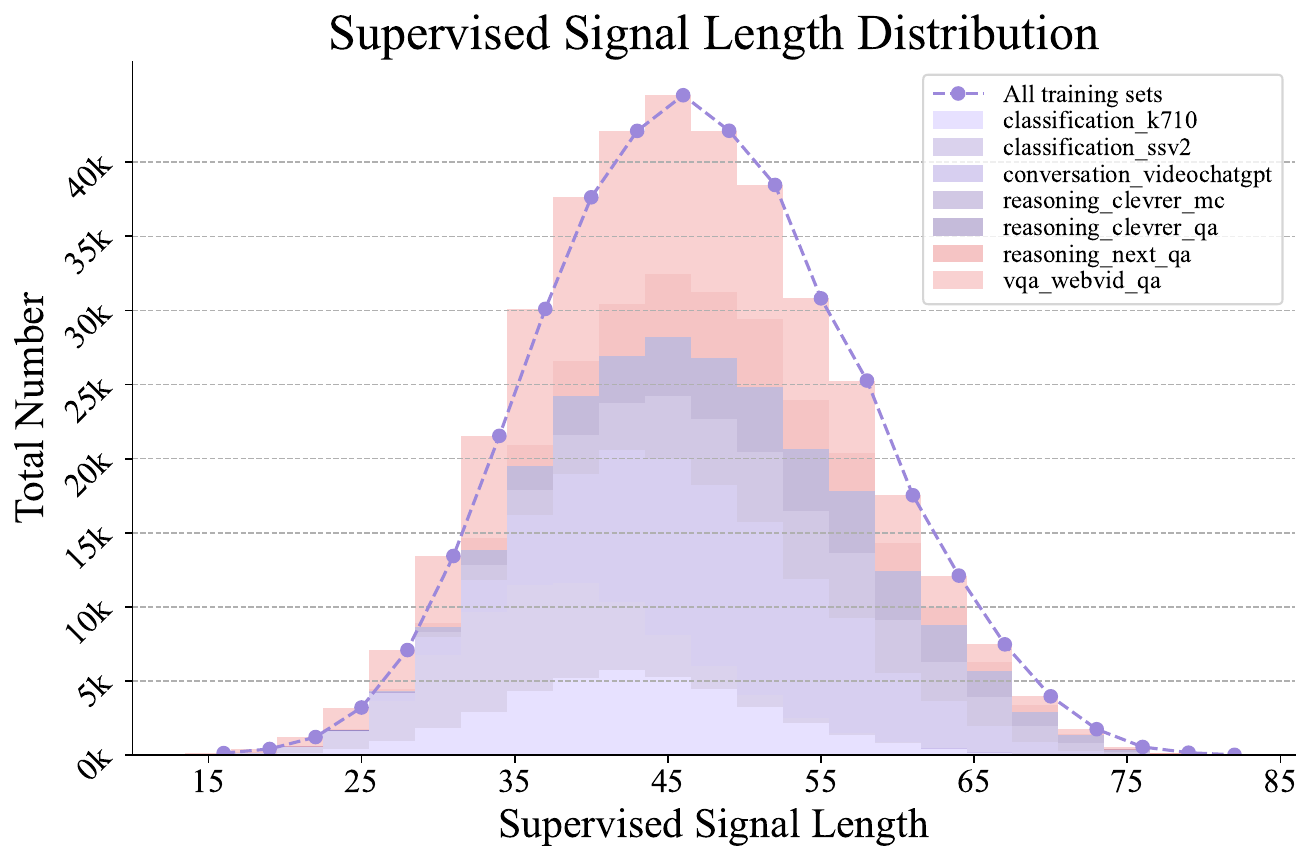}
        \caption{Distribution of supervision signal lengths.}
    \end{subfigure}
    \begin{subfigure}[b]{0.92\columnwidth}
        \centering
        \includegraphics[width=\textwidth]{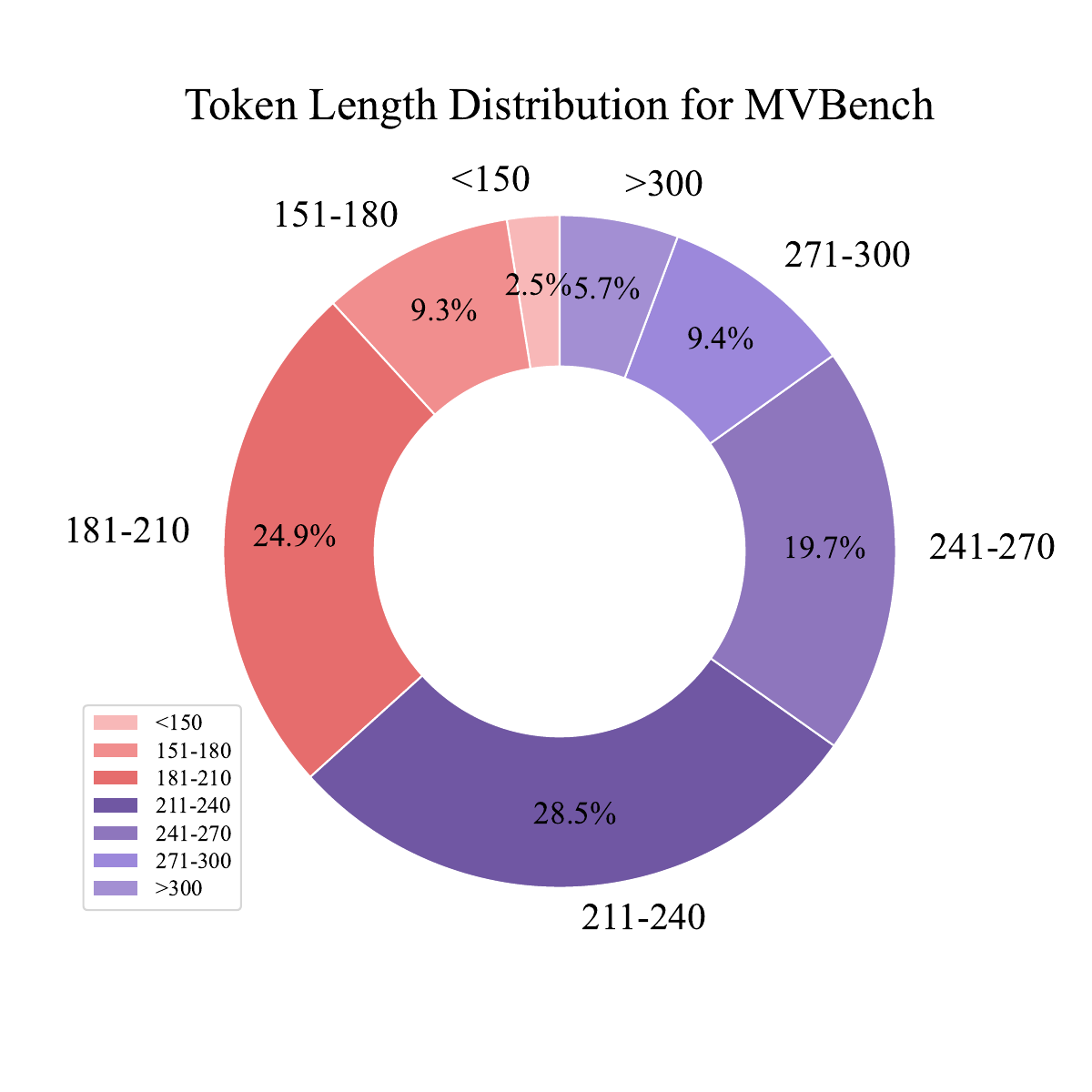}
        \vspace{-1.1cm}
        \caption{Distribution of compressed token lengths on MVBench.}
    \end{subfigure}
    \vspace{-0.2cm}
    \caption{Dynamic Token Length Distribution.}
    \label{fig:distribution}
\end{figure}

\noindent\textbf{Tangible demonstration of dynamic capabilities.}
To investigate the dynamic characteristics of our video compression method, we analyzed the length distributions of both the supervision signals during training and the compressed tokens in inference on the MVBench. Fig~\ref{fig:distribution} showcases  these distributions in two subplots. In subplot (a), we observe the distribution of supervision signal lengths for various video segments used in training, revealing insights into how the model learns to compress sequences of varying lengths. Moving to the inference phase, subplot (b) illustrates the distribution of the final compressed token lengths for complete videos from MVBench. This analysis not only highlights the overall compression effectiveness of \ourmodel but also sheds light on its adaptability to diverse video content.

\begin{figure*}[th]  
    \centering         
    \includegraphics[width=0.9\textwidth]{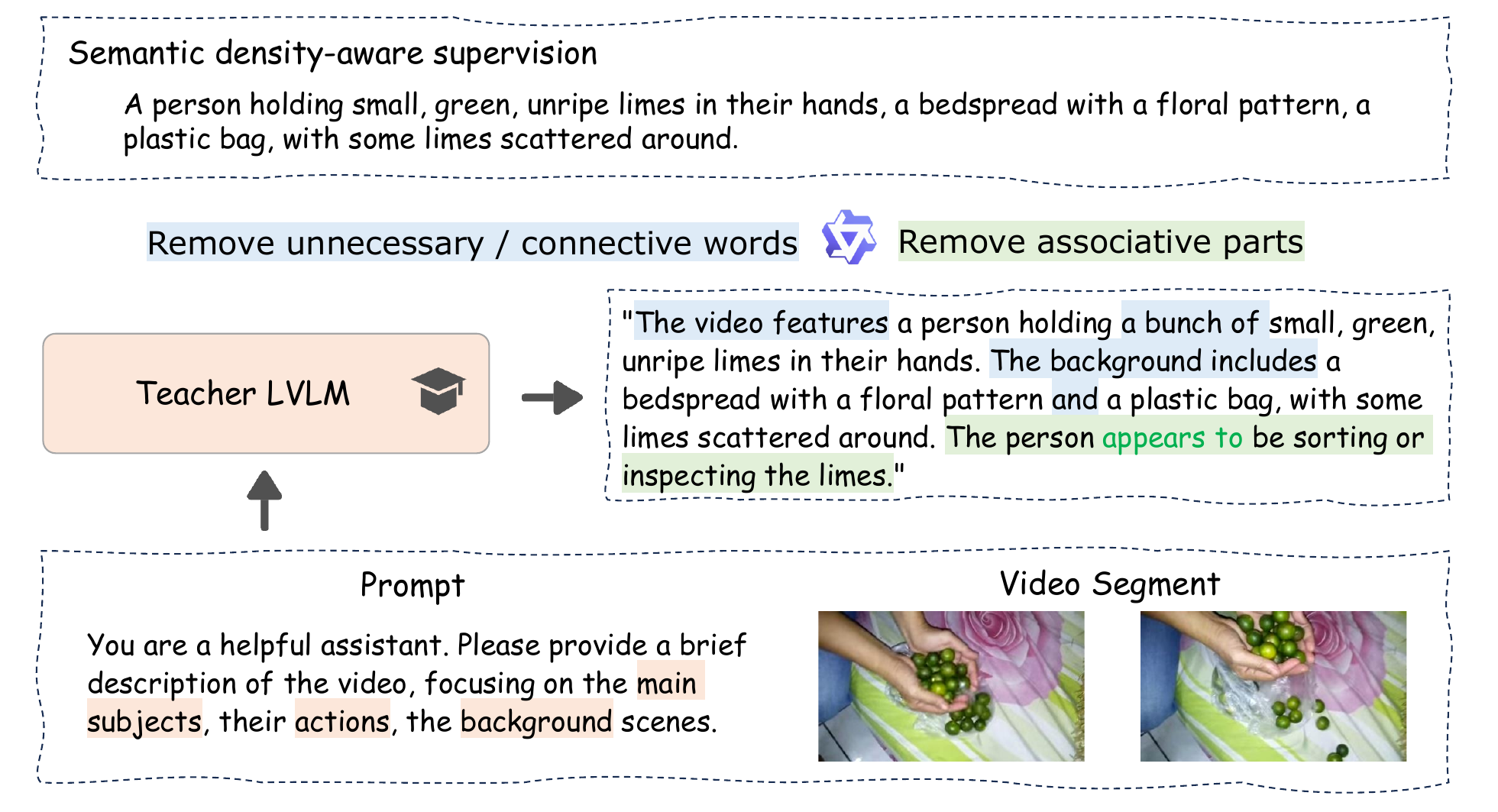}  
    \caption{ 
    The complete process of obtaining semantic density-aware supervision includes using a powerful LVLM as teacher to generate segment descriptions and a subsequent post-processing procedure.
    }
    \label{fig:process} 
\end{figure*}
\begin{table*}[!t]
    \centering
    \renewcommand{\arraystretch}{1.2}
    \setlength\tabcolsep{2pt}
    \resizebox{1.0\textwidth}{!}{
    \begin{tabular}{ccccccccccccccccccccccc}
        \toprule
        \multirow{2}{*}{\textbf{Method}} & \textbf{Efficiency} & \multicolumn{21}{c}{\textbf{Reference Metrics}}  \\
        \cmidrule(lr){2-2} \cmidrule(lr){3-23}
        &  Token Num.$\downarrow$& \textbf{AS} & \textbf{AP} & \textbf{AA} & \textbf{FA} & \textbf{UA} & \textbf{OE} & \textbf{OI} & \textbf{OS} & \textbf{MD} & \textbf{AL} & \textbf{ST} & \textbf{AC} & \textbf{MC} & \textbf{MA} & \textbf{SC} & \textbf{FP} & \textbf{CO} & \textbf{EN} & \textbf{ER} & \textbf{CI} & \cellcolor{gray!20}\textbf{Avg.} \\
        \midrule
        Pooling $2\times2$ & 3328 &  72.5 & 57.5 & 88.9 & 47.0 & 59.0 & 81.0 & 75.0 & 35.5 & 37.0 & 34.5 & 86.0 & 38.5 & 65.0 & 85.5 & 41.0 & 41.8 & 49.5 & 33.0 & 42.0 & 57.5 & \cellcolor{gray!20}55.37  \\
Pooling $4\times4$ & 832 &  67.5 & 54.0 & 73.7 & 44.0 & 57.0 & 73.5 & 70.5 & 35.0 & 26.5 & 35.0 & 85.5 & 36.5 & 54.5 & 78.0 & 40.0 & 40.5 & 43.0 & 34.0 & 40.0 & 52.5 & \cellcolor{gray!20}52.05 \\
Pooling $8\times8$ & 208  & 66.0 & 52.5 & 76.8 & 48.0 & 53.5 & 67.0 & 69.5 & 40.0 & 26.0 & 34.0 & 79.0 & 40.5 & 50.0 & 59.0 & 39.5 & 37.0 & 38.5 & 33.5 & 36.0 & 44.0 & \cellcolor{gray!20}49.50 \\
Pooling $16\times16$ & 80 & 57.5 & 45.0 & 69.7 & 44.0 & 49.5 & 49.5 & 68.5 & 33.0 & 19.5 & 28.0 & 80.0 & 38.0 & 47.0 & 49.0 & 39.0 & 34.5 & 33.0 & 32.0 & 35.5 & 36.0 & \cellcolor{gray!20}44.40 \\
        
        \midrule
        \midrule
         \textcolor{blue}{Oracle Performance} & \textcolor{blue}{--} &\textcolor{blue}{88.5}&\textcolor{blue}{74.0}&\textcolor{blue}{95.5}&\textcolor{blue}{63.0}&\textcolor{blue}{72.5}&\textcolor{blue}{96.5}&\textcolor{blue}{86.0}&\textcolor{blue}{67.5}&\textcolor{blue}{64.0}&\textcolor{blue}{60.0}&\textcolor{blue}{91.0}&\textcolor{blue}{49.0}&\textcolor{blue}{81.5}&\textcolor{blue}{96.5}&\textcolor{blue}{51.0}&\textcolor{blue}{61.5}&\textcolor{blue}{71.0}&\textcolor{blue}{50.0}&\textcolor{blue}{57.0}&\textcolor{blue}{72.0} & \cellcolor{gray!20}\textcolor{blue}{72.4} \\
         \textcolor{blue}{Oracle Tokens} & \textcolor{blue}{--} & \textcolor{blue}{355.4}&\textcolor{blue}{270.6}&\textcolor{blue}{405.9}&\textcolor{blue}{260.3}&\textcolor{blue}{256.7}&\textcolor{blue}{274.3}&\textcolor{blue}{233.4}&\textcolor{blue}{373.8}&\textcolor{blue}{757.8}&\textcolor{blue}{381.2}&\textcolor{blue}{156.5}&\textcolor{blue}{253.2}&\textcolor{blue}{507.9}&\textcolor{blue}{514.0}&\textcolor{blue}{211.4}&\textcolor{blue}{386.0}&\textcolor{blue}{497.4}&\textcolor{blue}{244.7}&\textcolor{blue}{263.5}&\textcolor{blue}{485.5} & \cellcolor{gray!20}\textcolor{blue}{354.48} \\
        \bottomrule
    \end{tabular}
    }
    \caption{
    A detailed examination of the performance comparison of pooling strategies with various compression rates on the entire MVBench benchmark. 
    \textcolor{blue}{Oracle} denotes the case where the best tradeoff between visual tokens and performance is picked. 
    Videos across different tasks have varying information loads, with the ideal token count differing significantly.
    }
\label{tab:full_pool_analysis}
\end{table*}

\begin{figure}[H]  
    \centering         
    \includegraphics[width=0.85\linewidth]{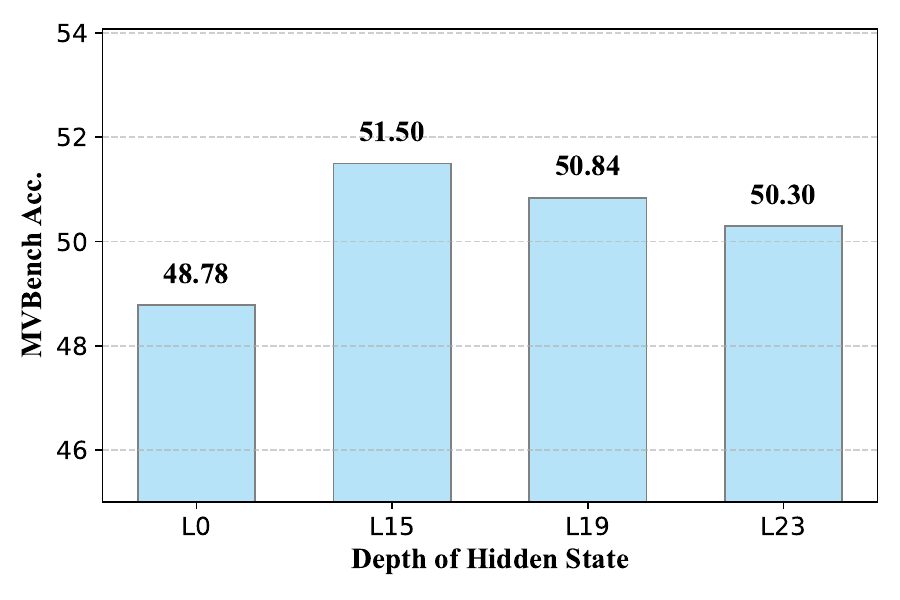}  
    \vspace{-5pt}
    \caption{
    Ablation of Hidden States Depth.
    }    
    \label{fig:ablation_llm} 
\end{figure}

\noindent\textbf{Ablation study on depth of hidden state.}
There is an interesting phenomenon that among the variable-length tokens generated by CapPruner, it is not the last layer's hidden states that perform the best as soft caption tokens. 
Figure \ref{fig:ablation_llm} illustrates that among the depth of hidden states, the zeroth layer performs the worst due to its weaker semantic information. Meanwhile, the middle layers exhibit slightly better performance than the last layer, possibly because representations that are too closely tied to the final classification task are more prone to overfitting, which may weaken their general representational capacity. In this ablation, we do not use BasePruner and fix the LLM as Qwen-2.5-1.5B.

\noindent\textbf{Effectiveness of semantic density-aware supervision.}
\renewcommand{\arraystretch}{1.2}
\begin{table}[!t]
    \centering
    \small
    \setlength\tabcolsep{5.0pt}
    \resizebox{0.95\linewidth}{!}{
        \begin{tabular}{c|c|c}
        \toprule
        \textbf{Base Caption Dataset} & \textbf{Semantic Supervision} &  \textbf{Accuracy}  \\
        \midrule
        {--} &  \ding{55} & 45.40\\
        \midrule
        \multirow{2}{*}{$COCO_{recap}$\cite{Li2024WhatIW}} &  \ding{55} & 46.80 (\darkGreen{$\uparrow$1.40}) \\
        ~ & \ding{52} & 49.98 (\darkGreen{$\uparrow$4.98})\\
        \midrule
        \multirow{2}{*}{$LLaVA_{recap}$\cite{liu2024llavanext}} & \ding{55} &  47.26 (\darkGreen{$\uparrow$1.86}) \\
        ~ & \ding{52} & 50.30 (\darkGreen{$\uparrow$4.90}) \\
        \bottomrule
        \end{tabular}
    }
    \caption{Ablation of the choice of base caption dataset and semantic density-aware supervision on MVBench.}
\label{tab:ablation_capdata}
\end{table}
To enhance CapPruner's sensitivity to visual information density, increased training with explicit supervision is essential. 
As shown in Table~\ref{tab:ablation_capdata}, CapPruner trained without high-quality vision-language pairs from the base caption dataset fails to produce compact and effective visual representations, resulting in poorer performance. 

Furthermore, naive caption supervision is inadequate and our semantic supervision is critical for achieving optimal results. For this ablation study, the deepest hidden state was chosen as the compressed representation.

\noindent\textbf{Generalizability of LangDC.} 
Table~\ref{tab:longvideo_results} below assess the generalizability of LangDC from two complementary perspectives. First, VSI-Bench~\cite{Yang_2025_CVPR} introduces a novel indoor-video benchmark, presenting scenes and configurations not seen during training. Remarkably, LangDC matches the baseline performance despite this unseen setting, demonstrating strong adaptability to new environments. Second, VideoMME-Long and LongVideoBench~\cite{wu2024longvideobench} assess the model’s capability to extract salient information from extended video sequences. LangDC maintains robust performance even without being explicitly trained on long-video data, indicating its ability to dynamically allocate visual tokens and capture key cues over long temporal spans. Together, these results highlight LangDC’s strong generalization across both unfamiliar indoor scenes and lengthy video content, underscoring its potential as a versatile video understanding framework.

\begin{table*}[htbp]
\centering
\adjustbox{max width=\textwidth}{ 
\begin{tabular}{l c *{6}{c}} 
\toprule
\multirow{2}{*}{Method} & \multirow{2}{*}{Token Num} & \multicolumn{3}{c}{VSI-Bench} & \multicolumn{2}{c}{VideoMME-long} & \multirow{2}{*}{LongVideoBench (val w/o subs)} \\
\cmidrule(lr){3-5} \cmidrule(lr){6-7} 
& & Object Appearance Order & Object Real Distance & Overall Multi-choice & w/o sub & w/ sub & \\
\midrule
VideoGPT+ & 3328 & 10.84 & \textbf{36.90} & 30.26 & 37.0 & 43.9 & 37.50 \\
LangDC    & \textbf{1068} & \textbf{14.24} & 36.06 & \textbf{30.82} & \textbf{38.9} & \textbf{46.4} & \textbf{43.83} \\
\bottomrule
\end{tabular}
}
\caption{Comparison of Methods on Various Video Benchmarks}
\label{tab:longvideo_results}
\end{table*}

\section{Implementation Details}
\noindent\textbf{Additional details for CapPruner pretraining.}
\label{sec:super}
To allow CapPruner to dynamically compress visual features, it is crucial to construct supervision signals of appropriate length for effective guidance. This process begins with a powerful LVLM that describes the scene. We selecte LLaVA-OneVision~\cite{liu2024llavanext} as the teacher model to articulate the subjects, actions, and background in the video. 
However, these descriptions are often overly verbose. To refine the descriptions, we utilized a large language model, Qwen2.5-7B~\cite{qwen2.5}, to eliminate unnecessary words, connectives, and speculative elements, resulting in semantic density-aware supervision tailored for specific segments, as shown in Fig~\ref{fig:process}.

\noindent\textbf{Additional details for instruction tuning set.}
Follow VideoGPT+~\cite{Maaz2024VideoGPT+}, supervised fine-tuning uses two distinct instruction-tuning datasets tailored for different task formats. 
For Multiple-choice VQA, the model is trained on the Kinetics-710~\cite{kay2017kinetics}, Something-Something-v2 ~\cite{goyal2017something}, conversations from VideoChat~\cite{2023videochat}, CLEVRER~\cite{yi2019clevrer}, VQA dataset from WebVid~\cite{bain2021frozen} and NExT-QA~\cite{xiao2021next} datasets, totaling approximately 330K single-turn conversations. For Open-ended VQA, the model is trained on VideoInstruct100K~\cite{Maaz2023VideoChatGPT}, VCG+ 112K~\cite{Maaz2024VideoGPT+}, VideoChat~\cite{2023videochat} conversation and caption data, and VQA from WebVid~\cite{bain2021frozen}, amounting to roughly 260K single-turn conversations.

\begin{figure*}[hbtp]  
    \centering         
    \includegraphics[width=0.98\textwidth]{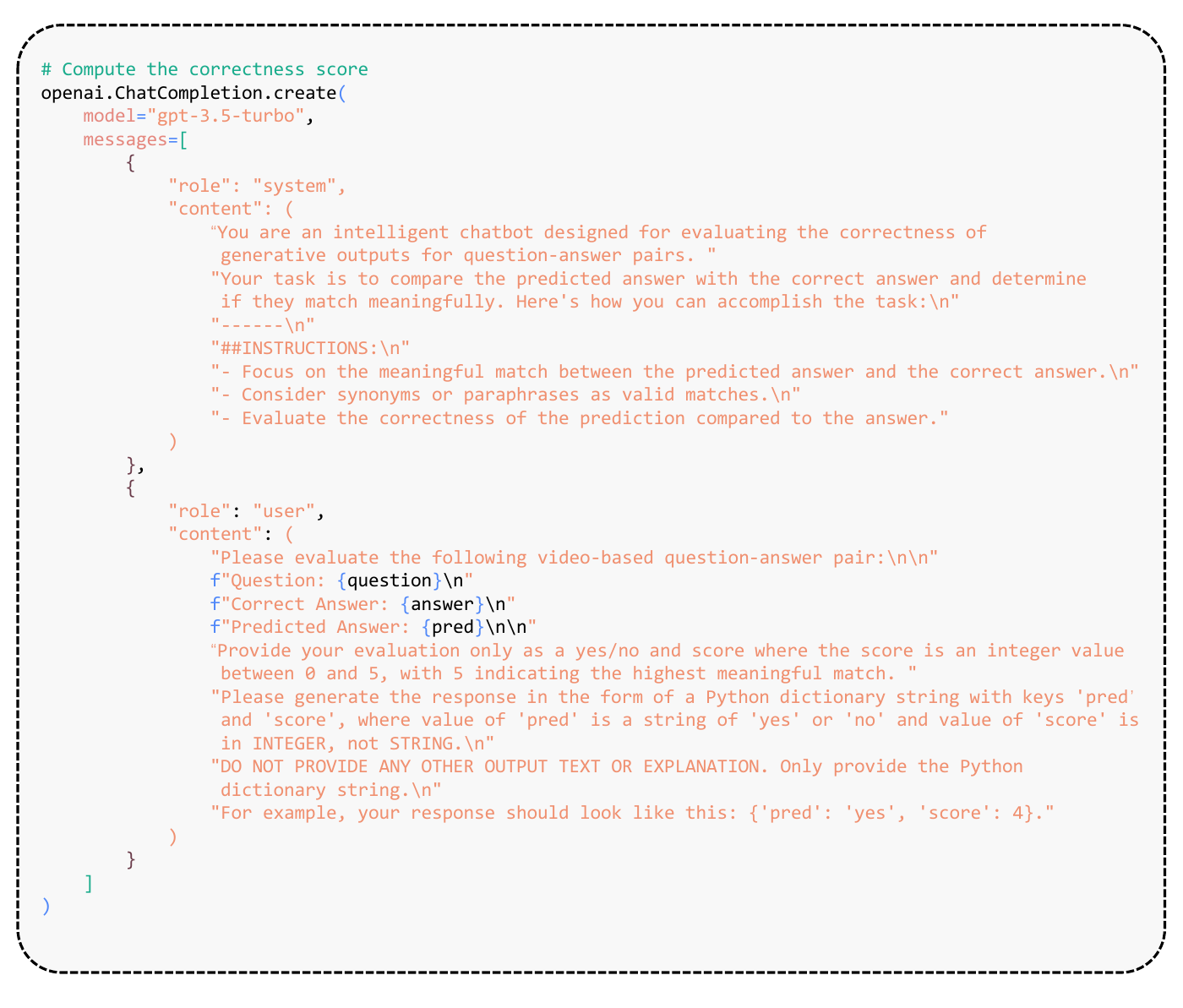}  
    \caption{ 
    Prompt for ChatGPT in LLM-Assisted Evaluation for the open-ended video question-answering task.
    }
    \label{fig:prompt} 
\end{figure*}

\noindent\textbf{Hyperparameter setting.}
\label{sec:hyper-param}
We report the detailed hyperparameter settings of \ourmodel in Tab.~\ref{tab:hyper}. During the training phase, each video is sampled into 16 frames and divided into 4 segments, with CapPruner compressing each segment to a maximum of 128 tokens, due to the longest supervision signal not exceeding 100 tokens. 

\label{sec:gpt-eval-varies}

\noindent\textbf{LLM-Assisted evaluation.} We utilize LLM-Assisted Evaluation for open-ended videoQA, following \cite{Maaz2023VideoChatGPT}. Each evaluation presents the LLM assistant (GPT-3.5) with the question, ground truth answer, and model prediction, prompting it to return a \textit{True} or \textit{False} judgement and a score (0-5). As depicted in Figure~\ref{fig:prompt}, this prompt uses roughly 250 tokens per question. Our baseline results for open-ended video question-answering are drawn from~\cite{Maaz2024VideoGPT+}.
\begin{table}[h]
\centering
\Large
\setlength{\tabcolsep}{8pt}
\renewcommand{\arraystretch}{1.3}
\resizebox{0.95\linewidth}{!}{
\begin{tabular}{ l | c}
\toprule
Description & Default Value \\
\midrule
total frame number & $16$ frames\\
segment number & $4$ segments\\
max compressed token number & $128$ tokens $\times 4$ segs\\
CapPruner hidden state layer & 15 \\
\bottomrule
\end{tabular}
}
\caption{Hyper-parameter settings of \ourmodel.}
\label{tab:hyper}
\end{table}

\section{Visualizations}
\label{sec:task_case}
\begin{figure*}[htbp]  
    \centering         
    \includegraphics[width=0.88\textwidth]{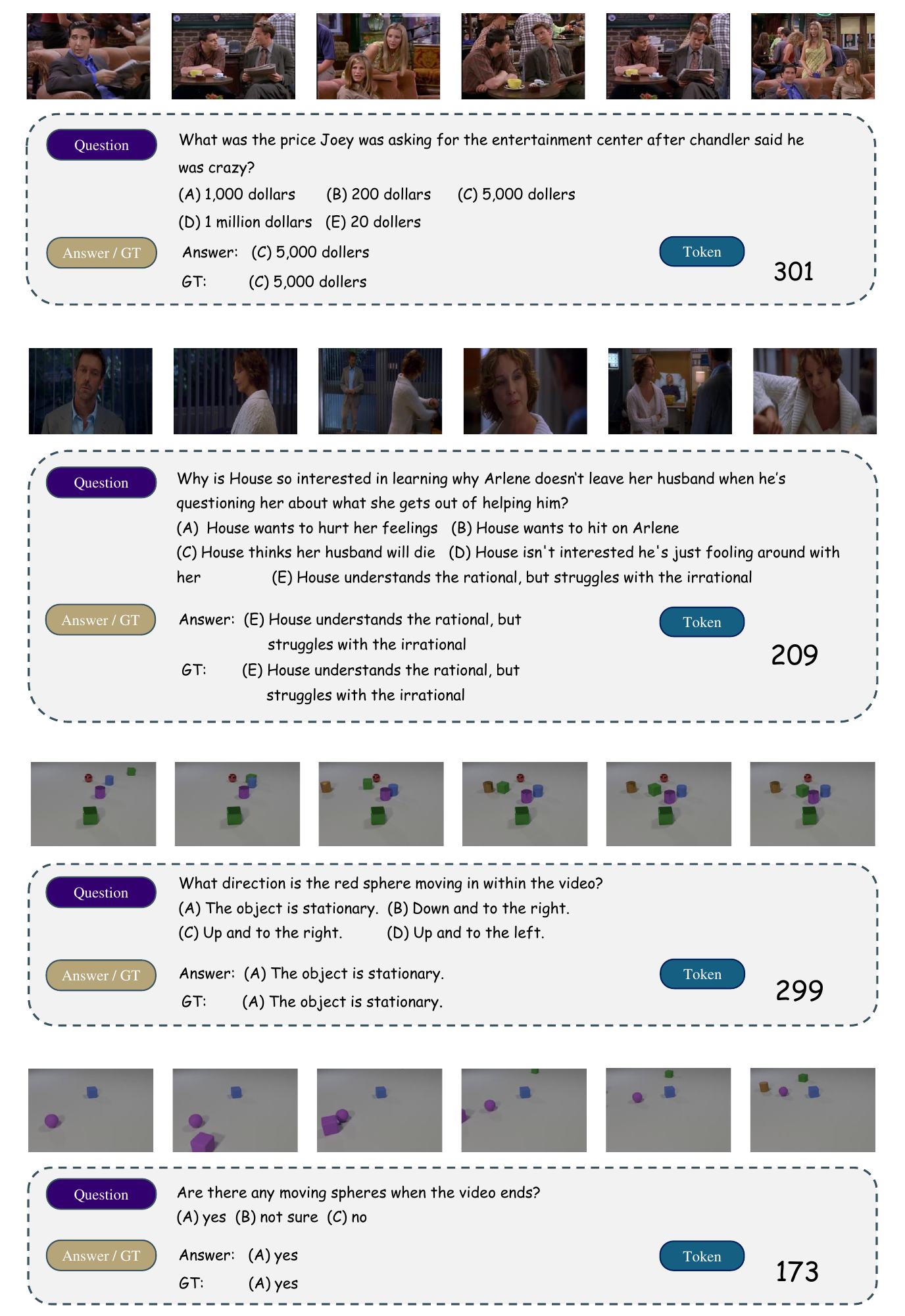}  
    \caption{ 
    More Cases.
    }
    \label{fig:case1} 
\end{figure*}

\begin{figure*}[htbp]  
    \centering         
    \includegraphics[width=0.88\textwidth]{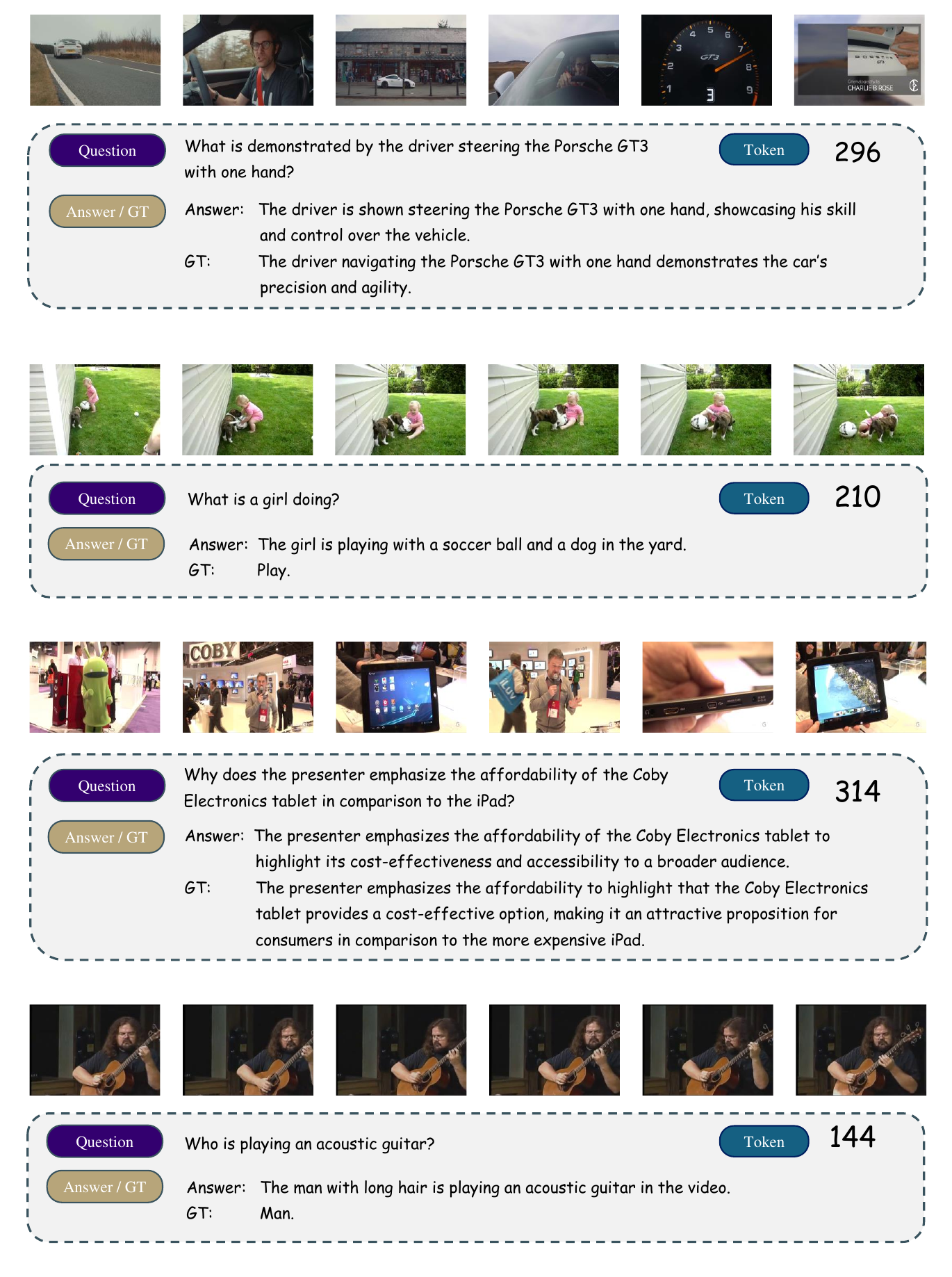}  
    \caption{ 
    More Cases.
    }
    \label{fig:case2} 
\end{figure*}

Figures \ref{fig:case1} and \ref{fig:case2} demonstrate the performance of \ourmodel and highlight how CapPruner adjusts the allocated token count based on the video content. 
These visualizations illustrate the overall token count after compression by CapPruner, along with video frames and question-answer pairs.
This effectively showcases the intelligence and adaptability of our compression scheme, as well as its resulting superior performance.

\end{document}